\documentclass[runningheads]{llncs}

\usepackage{graphicx}
\usepackage{hyperref}
\usepackage{xcolor}

\usepackage{caption}
\usepackage{subcaption}
\usepackage[T1]{fontenc}
\usepackage{placeins}

\begin{document}

\title{\textsc{ISImed}: A Framework for Self-Supervised Learning using Intrinsic Spatial Information in Medical Images}

\titlerunning{\textsc{ISImed}: Intrinsic Spatial Information in Medical Images}

 \author{Nabil Jabareen\inst{1}\orcidID{0000-0002-4170-9660} \and
 Dongsheng Yuan\inst{1}\orcidID{0009-0000-8819-3369} \and
 Sören Lukassen\inst{1} \orcidID{0000-0001-7045-6327}}

\authorrunning{Jabareen et al.}

 \institute{Berlin Institute of Health at Charité - Universitätsmedizin Berlin, \\
 Charitéplatz 1, 10117 Berlin, Germany \\
 \email{soeren.lukassen@charite.de}}

\maketitle        

\begin{abstract}
This paper demonstrates that spatial information can be used to learn interpretable representations in medical images using Self-Supervised Learning (SSL). 
Our proposed method, \textsc{ISImed}, is based on the observation that medical images exhibit a much lower variability among different images compared to classic data vision benchmarks.
By leveraging this resemblance of human body structures across multiple images, we establish a self-supervised objective that creates a latent representation capable of capturing its location in the physical realm.
More specifically, our method involves sampling image crops and creating a distance matrix that compares the learned representation vectors of all possible combinations of these crops to the true distance between them.
The intuition is, that the learned latent space is a positional encoding for a given image crop.
We hypothesize, that by learning these positional encodings, comprehensive image representations have to be generated. 
To test this hypothesis and evaluate our method, we compare our learned representation with two state-of-the-art SSL benchmarking methods on two publicly available medical imaging datasets.
We show that our method can efficiently learn representations that capture the underlying structure of the data and can be used to transfer to a downstream classification task.
\end{abstract}

\section{Introduction}
One major challenge of applying Deep Neural Networks (DNNs) to the medical domain, is posed by the necessity to have large amounts of labelled data to train supervised methods.
To label medical images, domain expertise is required.
Additionally, many medical imaging modalities are three dimensional, leading to even more time consuming, error prone and hence expensive labelling.
Self-supervised learning (SSL) is a rapidly growing area of machine learning that can be used to tackle these issues \cite{ssl-classification,ssl-in-medicine}.
The goal of SSL is to train models that can extract useful features and representations from data without the need for labeled training examples.
Many SSL methods rely on a joint embedding architecture similar to the Siamese network \cite{simclr,barlow}.
These methods use shared weights to generate similar embeddings to different views of an image.
A major challenge while training a SSL method is to prevent \textit{information collapse}, were the embedding for both views of an image become identical and constant.
To prevent this \textit{information collapse}, recent SSL architectures like simCLR \cite{simclr} and BarlowTwins \cite{barlow} use Contrastive Learning and Information Maximisation respectively. \newline
In medical imaging SSL is being successfully implemented in image classification \cite{ssl-classification}, segmentation \cite{ssl-segmentation} and other tasks like body part regression \cite{body-part-reg-21}.
 In \cite{ssl-segmentation} it was shown, that predicting anatomical positions as an objective for a SSL method can be used to boost the performance in cardiac MRI segmentation.
 Similarly, we use the anatomical position, i.e. the distance of sampled image patches, and show downstream classification capabilities on whole-body CT and brain MRI images.
 It has been shown that the accurate location of body parts can be used for content accumulation \cite{body-part-reg-21}. \newline
Our proposed approach can perform both, image representation learning and content accumulation by utilizing patch localization.
The upcoming section will provide an overview of our method and detail the benchmark experiments conducted to validate its effectiveness.

\section{Material \& Methods}
\subsection{Description of \textsc{ISImed}.}
\label{sec:desc_ISImed}
We propose a method for learning medical image representation using physical distance as a learning signal.
To train the model we randomly sample patches from a batch of images.
The physical $L_2$ distance between all pairs of these patches ($D_{physical}$) serves as our target.
Our method creates an embedding vector for every patch and finally the $L_2$ distance between every embedding pair ($D_{latent}$) is computed.
Our loss is then simply defined as:
\begin{equation}
\label{eq:loss}
    Loss = L_2(D_{physical}, D_{latent})
\end{equation}
The purpose of this objective function is, that the learned latent representations resemble the physical location of the encoded patches.
We hypothesize, that the learned features can be used for downstream transfer learning tasks as well as for content accumulation through body part regression. \newline
This method can be easily combined with other SSL methods (see section~\ref{sec:barlowdist}).
Additionally the $L_2$ distance in~(\ref{eq:loss}) could be replaced with any other distance-based regression loss.
The implementation of the Huber loss function is a potential optimization strategy for imposing greater penalties on distant patches.
However, due to the scope of this paper, further optimisations of this objective function will not be expounded upon further. 
Another advantage of our proposed method is that it works without the necessity of a joint embedding and does not need image distortions.
This reduction of hyperparameters makes optimization and training easier and computationally less expensive.

\subsubsection{Regularization of \textsc{ISImed}.}
\label{sec:barlowdist}
BarlowTwins can be seen as a information maximisation regularization of the latent representation \cite{barlow}.
\textsc{ISImed} however, is inspired by the idea to make the latent representation resemble physical space.
This can lead to an information collapse in \textsc{ISImed}, where most dimensions of the representation are non-informative.
Combining both methods can prevent the information collapse, while ensuring the latent representation remains a resemblance of the physical space.
Here we will describe the details of how we used BarlowTwins to regularize \textsc{ISImed} (\textsc{regISImed}). \newline
To combine the methods we trained the model as a joint embedding method just as in BarlowTwins \cite{barlow}.
A backbone network (as described in section \ref{sec:training}) generates a $1024$ latent representation for each view of the input batch.
A linear layer is added to the latent representation, producing a $512$ vector to which the \textsc{ISImed} loss is applied.
Another linear layer outputs a $2048$ vector to which the BarlowTwins loss is applied.
In order to balance the impact of two distinct loss functions, we introduced a hyperparameter denoted by $\lambda$, which is applied as a scalar coefficient to the BarlowTwins loss.
After analyzing the values of the losses for both BarlowTwins and \textsc{ISImed}, we determined that setting $\lambda$ to $10^3$ would result in roughly equivalent contributions from both losses towards the final loss.
Finally, the original latent representation of size $1024$ is used for all transfer tasks. \newline
Here we want to stress, that the linear layers are not to be considered as projector heads.
As shown in \cite{simclr}, for the model to benefit from a projector head, the projector head has to contain nonlinearities.

\subsection{Experimental setup and implementation details}
To benchmark our method, we trained on two different datasets and validated the performance on a downstream classification task.
All code will be made publicly available on GitHub (\url{https://github.com/NabJa/isimed}).
Additionally, all conducted experiments and models will be made available on Weights \& Biases (\url{https://wandb.ai/ag-lukassen/ISImed}).

\subsubsection{Training.}
\label{sec:training}
In our experiments all backbones are set to the DenseNet121 \cite{densenet} and trained for 50 epochs. 
As optimizer we choose the AdamW \cite{adamw} with an initial learning rate of $0.001$.
We applied an exponential learning rate decay with multiplicative factor of $0.9$.
Our overall batch size was set as a combination of the number of randomly sampled patches and the number of images these patches are sampled from.
In total we sampled $16$ patches from $64$ patients for every batch, resulting in a batch size of $1024$.
All models were implemented using MONAI and PyTorch and trained on NVIDIA A100 40GB GPUs.

\subsubsection{Benchmarking methods.}
As benchmarking methods we used simCLR \cite{simclr} and BarlowTwins \cite{barlow}.
Both methods sample from the same probability distribution of augmentations.
As augmentations we added random coarse dropout \cite{inpainting,cutout} and random coarse shuffling \cite{pixel-shuffle}.
The random coarse dropout was either applied on 6 cutouts (foreground-dropout) or 6 positions were kept unchanged, and the dropout was applied on the remaining voxels (background-dropout).
For the foreground-dropout the cutout patch size was uniformly sampled between $3\times3\times3$ voxels and $10\times10\times10$ voxels.
The patch size of the unchanged patches in the background-dropout mode was uniformly sampled between $3\times3\times3$ voxels and $21\times21\times21$ voxels.
Finally, the random coarse shuffling was applied with a probability of $80\%$. \newline
An important hyperparameter in simCLR is $\tau$, that weighs the implicit negative examples and positive examples for a given batch \cite{simclr}.
To find a good $\tau$ for our experiments we ran a random hyperparameter search, where $\tau$ was sampled from a log-uniform distribution between $0.05$ and $0.5$.
After conducting 29 trials we set $\tau$ to be $0.05$ for all experiments. \newline
BarlowTwins hyperparameter $\lambda$, which balances its invariance and redundancy reduction term \cite{barlow}, was optimized by conducting a random hyperparameter search with a log-uniform distribution ranging from $0.001$ to $0.1$.
After conducting 90 trials, we observed that the value of $\lambda$ did not demonstrate a significant impact on the validation loss within the specified range of sampling.
Therefore, we adopted the value of $\lambda = 0.005$, as suggested by the original publication \cite{barlow}. \newline
Apart from the mentioned hyperparameters and the augmentation strategy, the choice of the projector head size influences the models performance \cite{barlow}.
To make the models more comparable and reduce the number of parameters that have to be optimized, we did not add any projector head to any model.

\subsection{Datasets}
\label{sec:datasets}
We tested and benchmarked our method on two very different datasets with a minimal amount of preprocessing.
For both datasets, the image intensities were scaled using MONAIs \textit{ScaleIntensityRangePercentiles} using the $5\%$ and $95\%$ percentiles and $-1$ and $1$ as the target range.
All images were cropped to their foreground using MONAIs \textit{CropForeground} method.

\subsubsection{autoPET.}
The whole-body FDG-PET/CT dataset (autoPET) \cite{autopet} was obtained from the Cancer Imaging Archive (TCIA) \cite{tcia}.
From this dataset we only used the CT images and discarded all PET images.
Additionally, images with more than 400 axial image slices were filtered out to ensure a better image registration.
A rigid registration was applied to the datasets mean image using AirLab \cite{airlab}.
All images had a pixel spacing of $2mm\times2mm\times3mm$ (sagittal, coronal, axial).
Finally, the resulting 850 images were split into 515 training images, 172 validation images and 163 test images.
It was ensured that patients who underwent multiple scans were assigned to the same data partition. \newline
This dataset is publicly available and can be accessed under TCIA Restricted License Agreement. \footnote{\url{https://www.cancerimagingarchive.net/collection/fdg-pet-ct-lesions/}}
\subsubsection{BraTS.}
The Brain Tumor Segmentation Challenge 2022 data \cite{brats1,brats2,brats3} was obtained from Synapse.
For our experiments we only used the T1 modality.
All images had a pixel spacing of $1mm\times1mm\times1mm$.
The total of 1251 images, were randomly split into 750 training images, 251 validation images and 250 test images. \newline
This dataset is publicly available and can be downloaded at Synapse. 
\footnote{\url{https://www.synapse.org/Synapse:syn27046444/}}

\section{Results}
To evaluate our learned representations, we benchmarked our method in two ways.
In section~\ref{sec:result_encode} we show, that our learned representations capture physical space as hypothesised in section~\ref{sec:desc_ISImed}.
Second, we evaluate our learned representations on a downstream image classification task as in previous state-of-the-art SSL benchmarks \cite{simclr,benchmarking-ssl,barlow}.\newline

\subsection{Learned representations encode physical space}
\label{sec:result_encode}
\begin{figure}[ht]
    \centering
    \begin{subfigure}{0.49\textwidth}
        \centering
        \includegraphics[width=\textwidth]{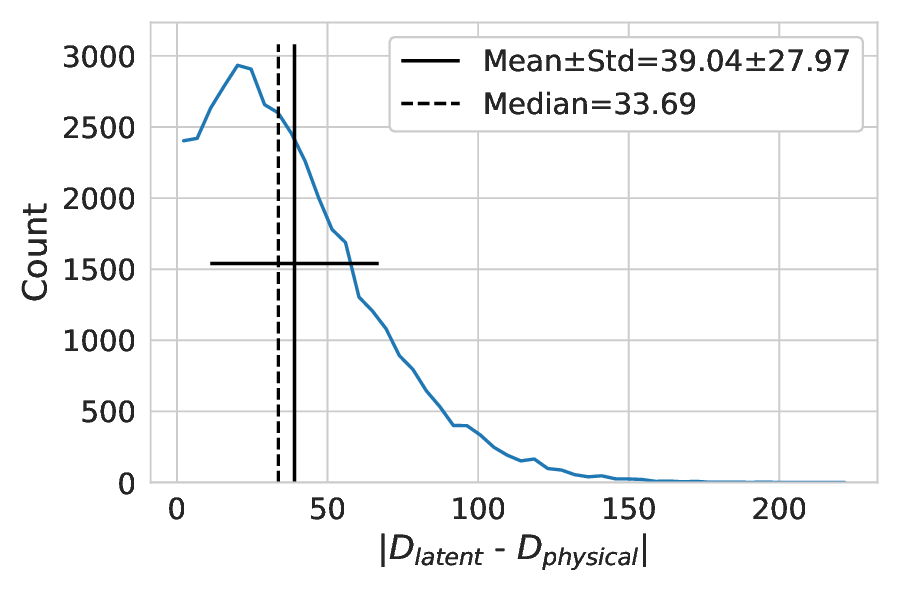}
        \caption{autoPET}
        \label{fig:pred_error_autopet}
    \end{subfigure}
    \begin{subfigure}{0.49\textwidth}
        \centering
        \includegraphics[width=\textwidth]{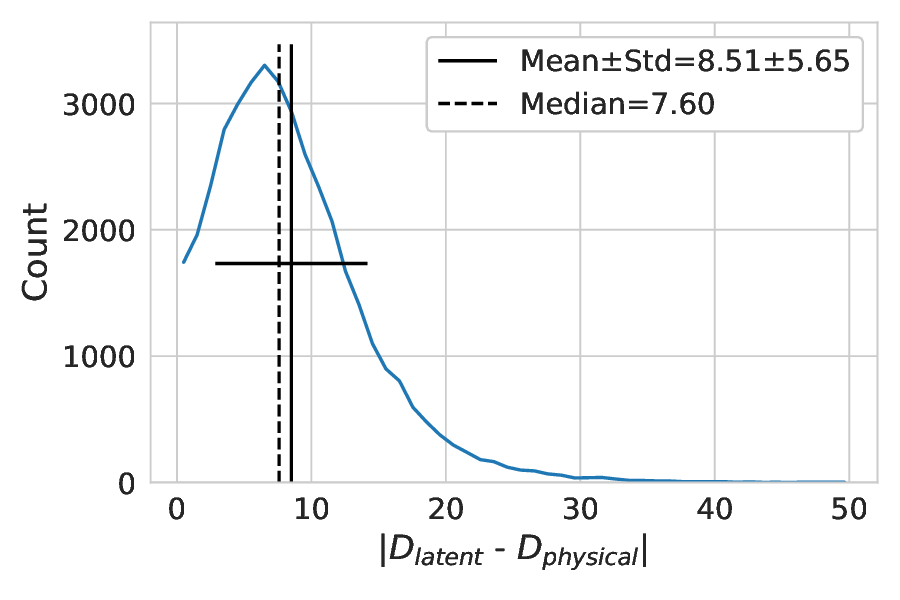}
        \caption{BraTS}
        \label{fig:pred_error_brats}
    \end{subfigure}
    \caption{Distribution of the absolute difference between the distance of the latent representations $D_{latent}$ and the true physical distance $D_{physical}$ of randomly sampled image patches.}
    \label{fig:pred_error}
\end{figure}
As shown in Fig.~\ref{fig:pred_error}, the distance between \textsc{ISImed}s learned representations ($D_{latent}$) closely match the true distance between the sampled patches ($D_{physical}$).
Since autoPET does have an anisotropic pixel spacing (see sec.~\ref{sec:training}), the distances have to be interpreted as voxel distances and can not be translated to millimeters.
In BraTS however, the pixel spacing is a $1mm\times1mm\times1mm$ isotropic pixel spacing and can therefore be interpreted as millimeters. 
Contrastive learning methods heavily depend on the sampling strategies for positive and negative pairs \cite{hardnegtaive} or on heavy augmentations and very large batch sizes \cite{simclr,barlow}.
Without the need of any augmentations and with random sampling, we show that our objective function (Eq. \ref{eq:loss}) can be learned in various datasets (see Fig.~\ref{fig:pred_error}).
\begin{figure}[ht]
    \centering
    \begin{subfigure}[b]{\textwidth}
        \centering
        \includegraphics[width=\columnwidth]{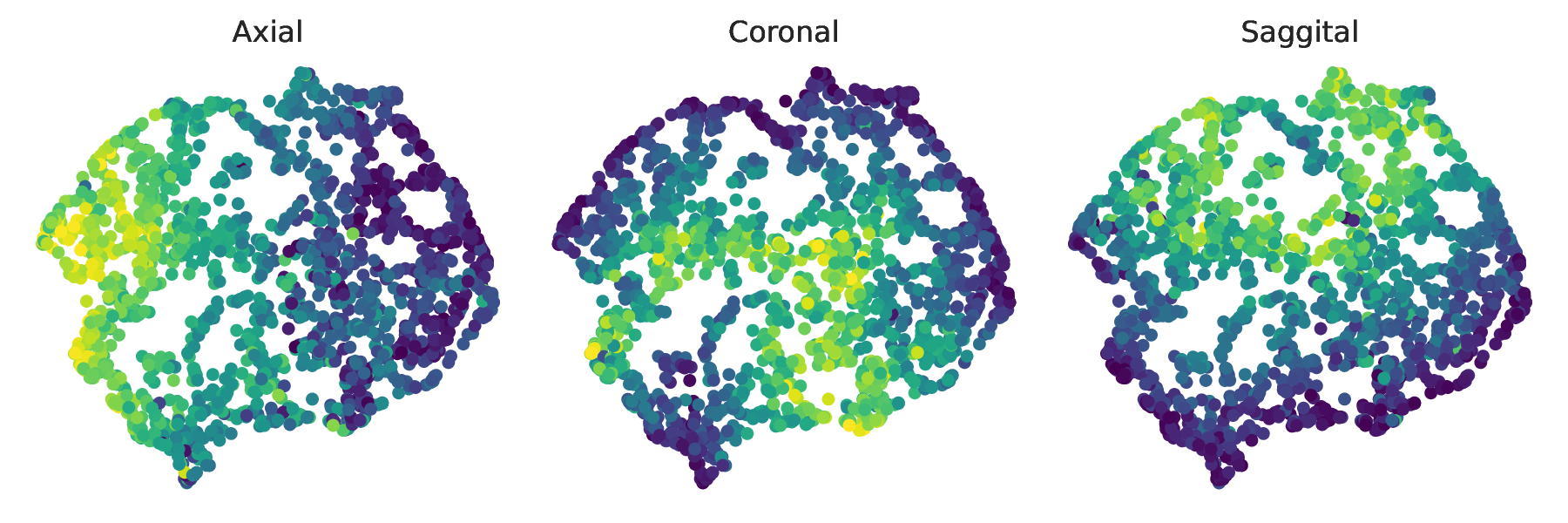}
        \caption{}
        \label{fig:umap}
    \end{subfigure}
    \begin{subfigure}[b]{\textwidth}
        \centering
        \includegraphics[width=\columnwidth]{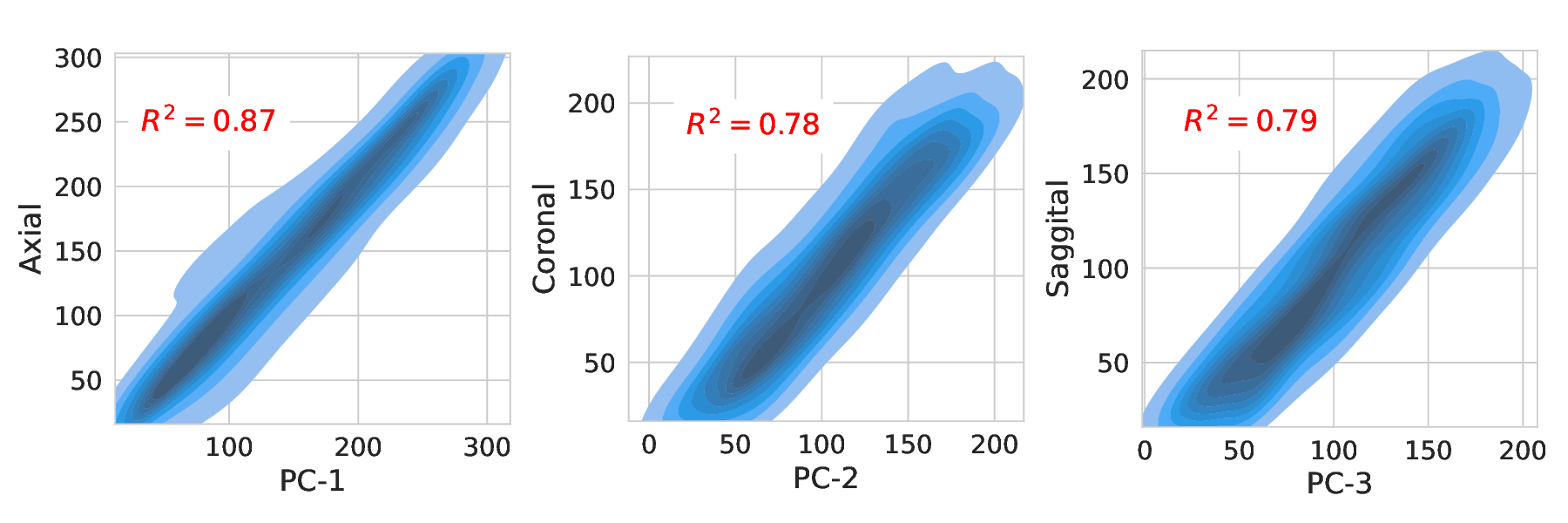}
        \caption{}
        \label{fig:body_part_regression}
    \end{subfigure}
    \caption{Learned latent representations reveal spatial location. In (a) a two dimensional UMAP is shown with the true spatial direction being indicated by the color. Note, that for all three directions the UMAP has a clear gradient, allowing to distinguish the patch location. In (b) the first three Principal Components of a PCA are shown alongside the true physical direction.}
\end{figure}
To further analyse the resulting latent representation of \textsc{ISImed}, we show our latent representation of patches from autoPET in a Uniform Manifold Approximation and Projection (UMAP) \cite{umap} (see Fig.~\ref{fig:umap}).
The color-coding is set to the axial, coronal and sagittal directions and it can be clearly seen, that all three dimension are captured in this latent representation.
In Fig.~\ref{fig:body_part_regression} we show the first three Principal Components (PCs) of the latent representations and their correlation with the spatial dimensions. \newline
Effective content navigation of medical images can be used to quickly and accurately locate and interpret patient information, potentially leading to improved diagnosis and treatment \cite{body-part-reg-21}.
Our results show that we are able to define a unified global coordinate system for a whole-body CT images (see Fig.~\ref{fig:body_part_regression}), as well as for brain MRI images (see supplementary material).
Compared to previous work on body part regression \cite{body-part-reg-21,body-part-reg-18} our method does not only predict the axial location of a given slice, but uses the three dimensional structure of the images and locates the patches in all three anatomical planes.

\subsection{Learned representations can be used for image classification}
\begin{figure}[ht]
    \centering
    \begin{subfigure}[b]{0.49\textwidth}
        \centering
        \includegraphics[width=\textwidth]{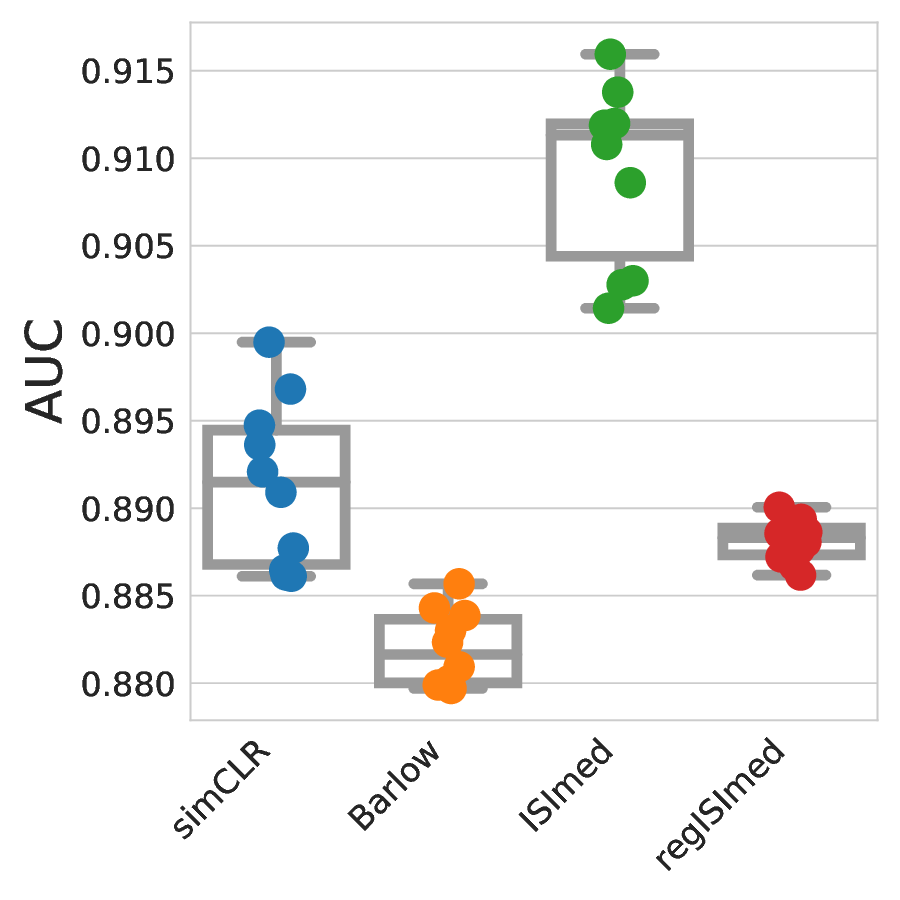}
        \caption{autoPET}
    \end{subfigure}
    \hfill
    \begin{subfigure}[b]{0.49\textwidth}
        \centering
        \includegraphics[width=\textwidth]{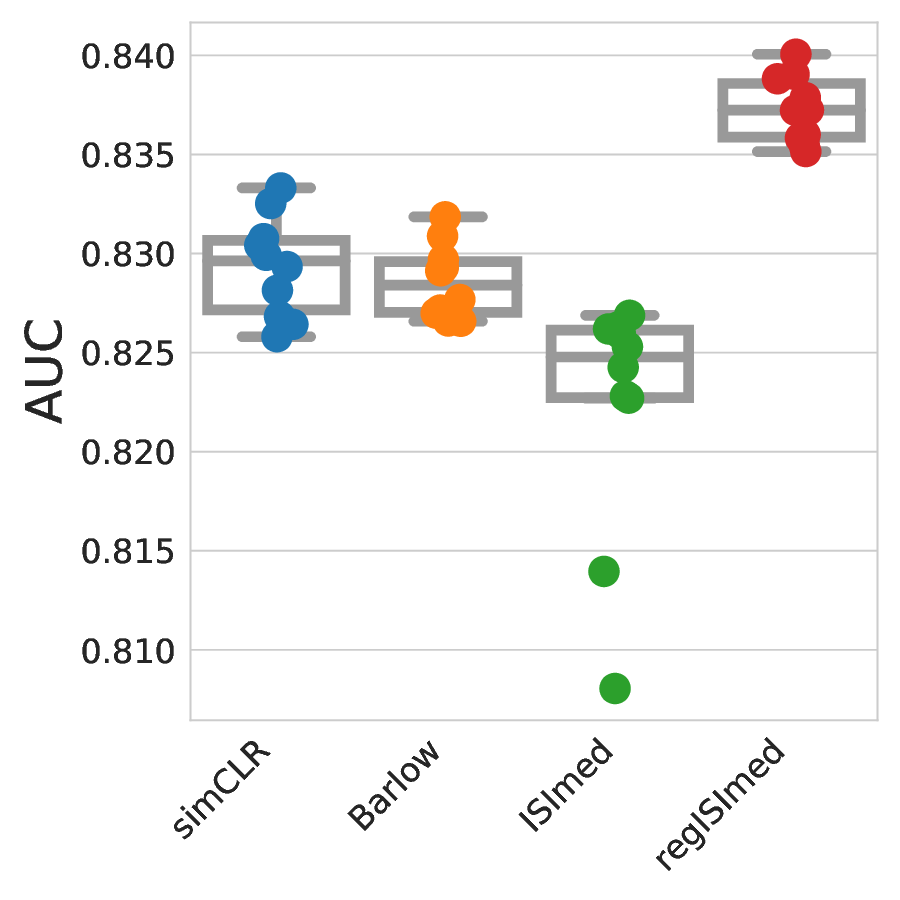}
        \caption{BraTS}
    \end{subfigure}
    \caption{AUC of downstream classification task. Shown is the performance of all folds from a 10-fold cross-validation. In the autoPET dataset \textsc{ISImed} significantly outperforms all other models (paired t-test $p<0.001$). In the BraTS dataset the combination of BarlowTwins and \textsc{ISImed} significantly outperforms all other models (paired t-test $p<0.001$).}
    \label{fig:downstream_auc}
\end{figure}
In this experiment we defined a downstream binary classification task.
For both datasets described in Section~\ref{sec:datasets} we sampled $32\times32\times32$ voxel patches.
The binary classification labels were then defined as whether the sampled patch contains any label.
In  the BraTS dataset these labels can be an edema, enhancing tumor or the necrotic core of the tumor.
In the autoPET dataset these labels can be malignant lymphoma, melanoma, or non-small cell lung cancer (NSCLC).
The samples were selected in a way that ensures the binary classification dataset is balanced, with an equal number of \textit{healthy} and \textit{anomalous} patches. \newline
All SSL models were trained on the training split.
Then a linear layer was fine tuned on the validation split without retraining the pre-trained SSL model.
Finally the models were benchmarked on the test set.
As shown in Table~\ref{tab:downstream}, \textsc{ISImed} significantly outperforms all other models in autoPET.
\begin{table}[ht]
    \begin{subtable}[ht]{\textwidth}
    \centering
    \resizebox{\columnwidth}{!}{%
        \begin{tabular}{lccccc}
            \hline
            Metric                                                      & AUC                        & Accuracy                   & F1-Score                   & Sensitivity                & Specificity                \\
            Model                                                       &                            &                            &                            &                            &                            \\ \hline
            simCLR                                                      & 0.891$\pm$0.005          & 0.821$\pm$0.005          & 0.796$\pm$0.003          & \textbf{0.823$\pm$0.006}          & 0.819$\pm$0.012          \\
            Barlow                                                      & 0.882$\pm$0.002          & 0.808$\pm$0.004          & 0.779$\pm$0.007          & 0.797$\pm$0.016          & 0.816$\pm$0.005          \\
            \textbf{\textsc{ISImed}}                                            & \textbf{0.909$\pm$0.005} & \textbf{0.837$\pm$0.006} & \textbf{0.809$\pm$0.008} & 0.818$\pm$0.012 & \textbf{0.850$\pm$0.003} \\
            \textsc{regISImed} & 0.888$\pm$0.001          & 0.813$\pm$0.001          & 0.783$\pm$0.001          & 0.798$\pm$0.001          & 0.824$\pm$0.001          \\ \hline
        \end{tabular}%
    }
    \caption{autoPET}
    \label{tab:downstream_autopet}
    \end{subtable}
    \hfill
    \begin{subtable}[ht]{\textwidth}
        \centering
        \resizebox{\columnwidth}{!}{%
            \begin{tabular}{lccccc}
                \hline
                Metric                                                               & AUC                        & Accuracy                   & F1-Score                   & Sensitivity                & Specificity                \\
                Model                                                                &                            &                            &                            &                            &                            \\ \hline
                simCLR                                                               & 0.829$\pm$0.002          & 0.712$\pm$0.003          & 0.691$\pm$0.007          & 0.646$\pm$0.009          & 0.778$\pm$0.004          \\
                Barlow                                                               & 0.829$\pm$0.002          & 0.731$\pm$0.002          & 0.723$\pm$0.003          & 0.697$\pm$0.003          & 0.765$\pm$0.002          \\
                \textsc{ISImed}                                                              & 0.822$\pm$0.006          & 0.724$\pm$0.043          & 0.685$\pm$0.036          & 0.638$\pm$0.039          & \textbf{0.799$\pm$0.054}          \\
                \textbf{\textsc{regISImed}} & \textbf{0.837$\pm$0.003} & \textbf{0.746$\pm$0.016} & \textbf{0.746$\pm$0.022} & \textbf{0.737$\pm$0.026} & 0.755$\pm$0.012 \\ \hline
            \end{tabular}%
        }

    \caption{BraTS}
    \label{tab:downstream_brats}
    \end{subtable}
    \caption{Downstream classification performance. Shown is the mean and standard deviation of the performance of a 10-fold cross validation.}
    \label{tab:downstream}
\end{table}
It is anticipated that \textsc{ISImed} would exhibit a high performance in the autoPET dataset due to the greater diversity of anatomical regions, in comparison to the BraTS dataset.
For whole-body imaging, such as that in the autoPET dataset, it is probable that the spatial arrangement of patches has a more significant impact than in smaller imaging modalities.
In the BraTS dataset, \textsc{regISImed} outperformed all other models to a statistically significant degree (paired t-test of AUC, $p<0.001$).
This finding suggests the hypothesis that, even in an organ-specific dataset such as BraTS, incorporating learned image representations that encode spatial location can be beneficial.

\section{Conclusion}
This paper introduces \textsc{ISImed}, a self-supervised representation learning model that establishes a direct relationship between the generated representations and physical space.
Notably, our approach offers an effortless implementation, without any additional hyperparameters to optimize, and seamless integration with other methods.
Most importantly, we demonstrate that our learned representations, endowed with intrinsic medical image information, achieve outstanding performance on downstream classification tasks across two diverse datasets. \newline
A limitation of our model is the susceptibility to \textit{information collapse}.
We have shown, that using regularization methods like BarlowTwins can mitigate this problem.\newline
Finally, our findings firmly establish the case for further exploration of intrinsic information in medical images, paving the way for enhancing self-supervised learning methods in medical imaging.

\newpage

\bibliographystyle{splncs04}
\bibliography{main.bib}

\newpage
\section{Supplementary Material}
\begin{figure}[ht]
    \centering
    \begin{subfigure}[b]{\textwidth}
        \centering
        \includegraphics[width=\columnwidth]{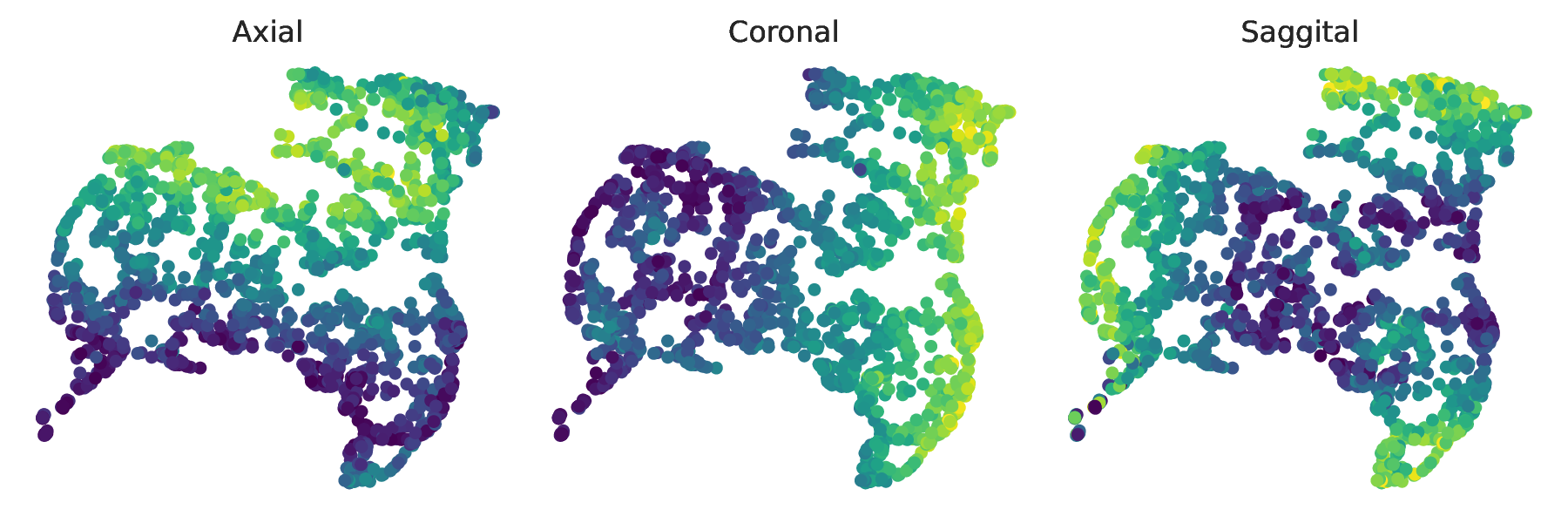}
        \caption{}
        \label{fig:umap_brats}
    \end{subfigure}
    \begin{subfigure}[b]{\textwidth}
        \centering
        \includegraphics[width=\columnwidth]{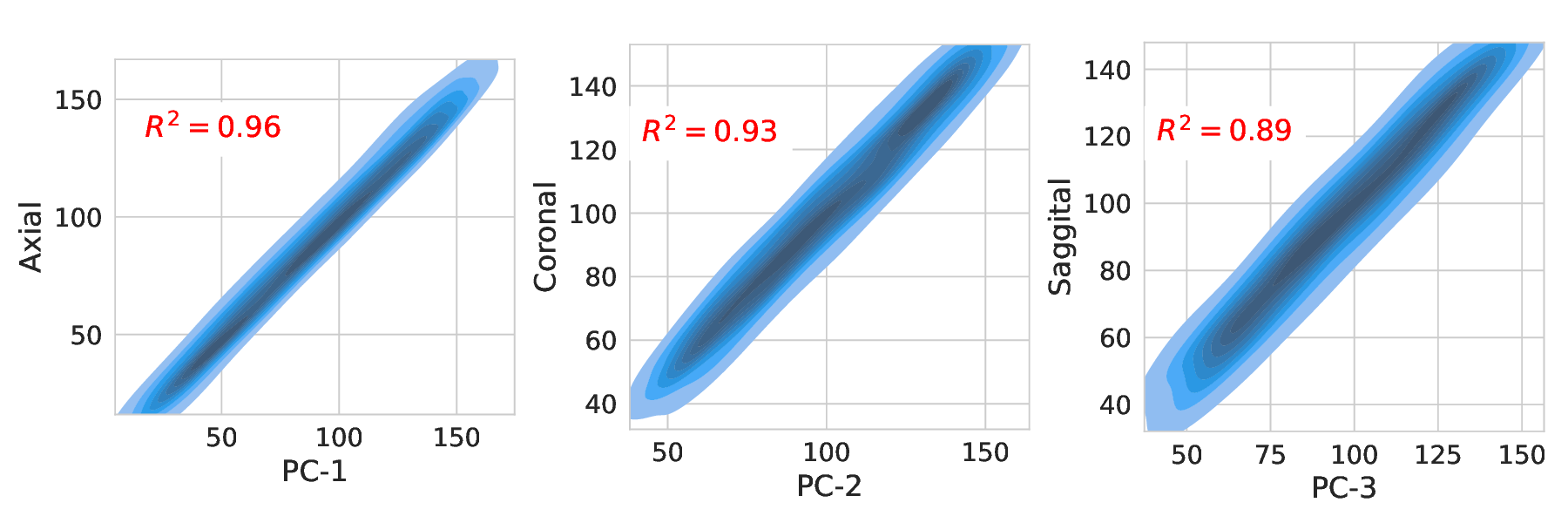}
        \caption{}
        \label{fig:body_part_regression_brats}
    \end{subfigure}
    \caption{Learned latent representations reveal spatial location in the BraTS dataset. In (a) a two dimensional UMAP is shown with the true spatial direction being indicated by the color. Note, that for all three directions the UMAP has a clear gradient, allowing to distinguish the patch location. In (b) the first three Principal Components of a PCA are shown alongside the true physical direction.}
\end{figure}

\end{document}